# Valuation Networks and Conditional Independence

**Prakash P. Shenoy**
School of Business
University of Kansas
Summerfield Hall
Lawrence, KS 66045-2003, USA
pshenoy@ukanvm.cc.ukans.edu

## Abstract

Valuation networks have been proposed as graphical representations of valuation-based systems (VBSs). The VBS framework is able to capture many uncertainty calculi including probability theory, Dempster-Shafer's belief-function theory, Spohn's epistemic belief theory, and Zadeh's possibility theory. In this paper, we show how valuation networks encode conditional independence relations. For the probabilistic case, the class of probability models encoded by valuation networks includes undirected graph models, directed acyclic graph models, directed balloon graph models, and recursive causal graph models.

## 1 INTRODUCTION

Recently, we proposed valuation networks as a graphical representation of valuation-based systems [Shenoy 1989, 1992a]. The axiomatic framework of valuation-based systems (VBS) is able to represent many different uncertainty calculi such as probability theory [Shenoy 1992a], Dempster-Shafer's belief-function theory [Shenoy 1993], Spohn's epistemic belief theory [Shenoy 1991a, 1992a], and Zadeh's possibility theory [Shenoy 1992b]. In this paper, we explore the use of valuation networks for representing conditional independence relations in probability theory and in other uncertainty theories that fit in the VBS framework.

Conditional independence has been widely studied in probability and statistics [see, for example, Dawid 1979, Spohn 1980, Lauritzen 1989, Pearl 1988, and Smith 1989]. Pearl and Paz [1987] have stated some basic properties of the conditional independence relation. (These properties are similar to those stated first by Dawid [1979] for probabilistic conditional independence, those stated by Spohn [1980] for causal independence, and those stated by Smith [1989] for generalized conditional independence.) Pearl and Paz call these properties 'graphoid axioms,' and they call any ternary relation that satisfies these properties a 'graphoid.' The graphoid axioms are satisfied not only by conditional independence in probability theory, but also by vertex separation in undirected graphs (hence the term graphoids) [Pearl and Paz 1987], by d-separation in directed acyclic graphs [Verma and Pearl 1990], by partial correlation [Pearl and Paz 1987], by embedded multi-valued dependency models in relational databases [Fagin 1977], by conditional independence in Spohn's theory of epistemic beliefs [Spohn 1988, Hunter 1991], and by qualitative conditional independence [Shafer, Shenoy and Mellouli 1987]. Shenoy [1991b, 1992c] has defined conditional independence in VBSs and shown that it satisfies the graphoid axioms. Thus the graphoid axioms are also satisfied by the conditional independence relations in all uncertainty theories that fit in the VBS framework including Dempster-Shafer's belief-function theory and Zadeh's possibility theory.

The use of undirected graphs and the use of directed acyclic graphs to represent conditional independence relations in probability theory have been extensively studied [see, for example, Darroch, Lauritzen and Speed 1980, Lauritzen 1989a,b, Wermuth and Lauritzen 1983, Kiiveri, Speed and Carlin 1984, Pearl and Paz 1987, Pearl, Geiger and Verma 1990, Lauritzen and Wermuth 1989, Frydenberg 1989, and Wermuth and Lauritzen 1990]. The use of graphs to represent conditional independence relations is useful since an exponential number of conditional independence statements can be represented by a graph with a polynomial number of vertices.

In undirected graphs (UGs), vertices represent variables, and edges between variables represent dependencies in the following sense. Suppose $a$, $b$, and $c$ are disjoint subsets of variables. The conditional independence statement '$a$ is conditionally independent of $b$ given $c$,' denoted by $a \perp b \mid c$, is represented in an UG if every path from a variable in $a$ to a variable in $b$ contains a variable in $c$, i.e., if $c$ is a *cut-set* separating $a$ and $b$. One can also represent a conditional independence relation by a set of UGs [Paz 1987]. A conditional independence relation is represented by a set of UGs if each independence statement in the relation is represented in one of the UGs in the set. In general, one may not be able to represent a conditional independence relation that holds in a probability distribution by one UG. Some probability distributions may require an exponential number of UGs to represent the conditional independence relation that holds in it [Verma 1987].

In directed acyclic graphs (DAGs), vertices represent variables, and arcs represent dependencies in the following sense. Pearl [1988] has defined d-separation of two sets of variables by a third. Suppose $a$, $b$, and $c$ are disjoint subsets of variables. We say $c$ d-separates $a$ and $b$ iff there is no path from a variable in $a$ to a variable in $b$ along which (1) every vertex with an outgoing arc is not in c, and (2) every vertex with incoming arcs is either in $c$ or has a descendant in $c$. The definition of d-separation takes into account the direction of the arcs in à DAG. The conditional independence statement $a \perp b \mid c$ is represented in a DAG if $c$ d-separates $a$ and $b$. One can also represent conditional independence relations by a set of DAGs [Geiger 1987]. A conditional independence relation is represented by a set of DAGs if it is represented in one of the DAGs in the set. As in the case of UGs, one may not be able to represent a conditional independence relation that holds in a probability distribution by one DAG. Some probability distributions may require an exponential number of DAGs to represent the conditional independence relations that hold in it [Verma 1987].

Shafer [1993a] has defined directed balloon graphs (DBGs) that generalize DAGs. A DBG includes a partition of the set of all variables. Each element of the partition is called a balloon. Each balloon has a set of variables as its parents. The parents of a balloon are shown by directed arcs pointing to the balloon. A DBG is acyclic in the same sense that DAGs are acyclic. A DBG implies a probability model consisting of a conditional for each balloon given its parents. A DAG may be considered as a DBG in which each balloon is a singleton subset. Independence properties of DBGs are studied in Shafer [1993b].

UGs and DAGs represent conditional independence relations in fundamentally different ways. There are UGs such that the conditional independence relation represented in an UG cannot be represented by one DAG. And there are DAGs such that the conditional independence relation represented in a DAG cannot be represented by one UG. In fact, Ur and Paz [1991] have shown that there is an UG such that to represent the conditional independence relation in it requires an exponential number of DAGs. And there is a DAG such that to represent the conditional independence relation in it requires an exponential number of UGs.

In valuation networks (VNs), there are two types of vertices. One set of vertices represents variables, and the other set represents valuations. Valuations are functions defined on variables. In probability theory, for example, a valuation is a factor of the joint probability distribution. In VNs, there are edges only between variables and valuations. There is an edge between a variable and a valuation if and only if the variable is in the domain of the valuation. If a valuation is a conditional for r given t, then we represent this by making the edges between the conditional and variables in r directed (pointed toward the variables). (Conditionals are defined in Section 2 and correspond to conditional probability distributions in probability theory.) Thus VNs explicitly depict a factorization of the joint valuation. Since there is a one-to-one correspondence between a factorization of the joint valuation and the conditional independence relation that holds in it, VNs also explicitly represent conditional independence relations.

The class of probability models included by VNs include UGs, DAGs and DBGs. Given a UG, there is a corresponding VN such that all conditional independence statements represented in the UG are represented in the VN. Given a DAG, there is a corresponding VN such that all conditional independence statements represented in the DAG are represented in the corresponding VN. And given a DBG, there is a corresponding VN such that all conditional independence statements represented in the DBG are represented in the corresponding VN.

Besides UGs, DAGs, and DBGs, there are other graphical models of probability distributions. Kiiveri, Speed, and Carlin [1984] have defined recursive causal graphs (RCGs) that generalize DAGs and UGs. Recursive causal graphs have two components, an UG on one subset of variables (exogenous), and a DAG on another subset of variables (endogenous). Given a RCG, there is a corresponding VN such that all conditional independence statements represented in the RCG are represented in the VN.

Lauritzen and Wermuth [1989] and Wermuth and Lauritzen [1990] have defined chain graphs that generalize recursive causal graphs. Conditional independence properties of chain graphs have been studied by Frydenberg [1990]. It is not clear to this author whether VNs include the class of probability models captured by chain graphs.

Jirousek [1991] has defined decision tree models of probability distributions. These models are particularly expressive for asymmetric conditional independence relations, i.e., relations that only hold for some configurations of the given variables, and not true for others. VNs, as defined here, do not include the class of models captured by decision trees.

Heckerman [1990] has defined similarity networks as a tool for knowledge acquisition. Like Jirousek's decision tree models, similarity networks allow representations of asymmetric conditional independence relations. VNs, as defined here, do not include the class of models captured by similarity networks.

An outline of this paper is as follows. Section 2 sketches the VBS framework and the general definition of conditional independence. The definition of conditional independence in VBS is a generalization of the definition of conditional independence in probability theory. Most of the material in this section is a summary of [Shenoy 1991b, 1992c]. Section 3 describes the valuation network representation and shows how conditional independence relations are encoded in valuation networks. Section 4 compares VNs to UGs, DAGs, DBGs, and RCGs. Finally, Section 5 contains some concluding remarks.

## 2 VBSs AND CONDITIONAL INDEPENDENCE

In this section, we briefly sketch the axiomatic framework of valuation-based systems (VBSs). Details of the axiomatic framework can be found in [Shenoy 1991b, 1992c].

In the VBS framework, we represent knowledge by entities called variables and valuations. We infer conditional independence statements using three operations called combination, marginalization, and removal. We use these operations on valuations.

**Variables.** We assume there is a finite set $\mathfrak{X}$ whose elements are called *variables*. Variables are denoted by upper-case Latin alphabets, X, Y, Z, etc. Subsets of $\mathfrak{X}$ are denoted by lower-case Latin alphabets, r, s, t, etc.

**Valuations.** For each $s \subseteq \mathfrak{X}$, there is a set $\mathcal{V}_s$. We call the elements of $\mathcal{V}_s$ *valuations for s*. Let $\mathcal{V}$ denote $\cup \{\mathcal{V}_s \mid s \subseteq \mathfrak{X}\}$, the set of all *valuations*. If $\sigma \in \mathcal{V}_s$, then we say *s is the domain of* $\sigma$. Valuations are denoted by lower-case Greek alphabets, $\rho$, $\sigma$, $\tau$, etc.

Valuations are primitives in our abstract framework and, as such, require no definition. But as we shall see shortly, they are objects that can be combined, marginalized, and removed. Intuitively, a valuation for s represents some knowledge about variables in s.

**Zero Valuations.** For each $s \subseteq \mathfrak{X}$, there is at most one valuation $\zeta_s \in \mathcal{V}_s$ called *the zero valuation for s*. Let $\mathcal{Z}$ denote $\{\zeta_s \mid s \subseteq \mathfrak{X}\}$, the set of all *zero valuations*. We call valuations in $\mathcal{V} - \mathcal{Z}$ *nonzero valuations*.

Intuitively, a zero valuation represents knowledge that is internally inconsistent, i.e., knowledge that is a contradiction, or knowledge whose truth value is always false. The concept of zero valuations is important in the theory of consistent knowledge-based systems [Shenoy 1990b].

**Proper Valuations.** For each $s \subseteq \mathfrak{X}$, there is a subset $\mathcal{P}_s$ of $\mathcal{V}_s - \{\zeta_s\}$. We call the elements of $\mathcal{P}_s$ *proper valuations for s*. Let $\mathcal{P}$ denote $\cup \{\mathcal{P}_s \mid s \subseteq \mathfrak{X}\}$, the set of all *proper valuations*. Intuitively, a proper valuation represents knowledge that is partially coherent. By coherent knowledge, we mean knowledge that has well-defined semantics.

**Normal Valuations.** For each $s \subseteq \mathfrak{X}$, there is another subset $\mathfrak{N}_s$ of $\mathcal{V}_s - \{\zeta_s\}$. We call the elements of $\mathfrak{N}_s$ *normal valuations for s*. Let $\mathfrak{N}$ denote $\cup \{\mathfrak{N}_s \mid s \subseteq \mathfrak{X}\}$, the set of all *normal valuations*. Intuitively, a normal valuation represents knowledge that is also partially coherent, but in a sense that is different from proper valuations.

We call the elements of $\mathcal{P} \cap \mathfrak{N}$ *proper normal valuations*. Intuitively, a proper normal valuation represents knowledge that is completely coherent, i.e., knowledge that has well-defined semantics.

**Combination.** We assume there is a mapping $\oplus : \mathcal{V} \times \mathcal{V} \to \mathfrak{N} \cup \mathcal{Z}$, called *combination*, such that if $\rho \in \mathcal{V}_r$ and $\sigma \in \mathcal{V}_s$, then $\rho \oplus \sigma \in \mathcal{V}_{r \cup s}$. Also we assume that combination is associative and commutative. Finally, suppose zero valuations exist, and suppose $\sigma \in \mathcal{V}_s$. Then we assume that $\zeta_r \oplus \sigma = \zeta_{r \cup s}$.

Intuitively, combination corresponds to aggregation of knowledge. If $\rho$ and $\sigma$ are valuations for r and s representing knowledge about variables in r and s, respectively, then $\rho \oplus \sigma$ represents the aggregated knowledge about variables in $r \cup s$.

It follows from the definition of combination that the set $\mathfrak{N}_s \cup \{\zeta_s\}$ together with the combination operation $\oplus$ is a commutative semigroup.

**Identity Valuations.** We assume that for each $s \subseteq \mathfrak{X}$, the commutative semigroup $\mathfrak{N}_s \cup \{\zeta_s\}$ has an identity denoted by $\iota_s$. In other words, there exists $\iota_s \in \mathfrak{N}_s \cup \{\zeta_s\}$ such that for each $\sigma \in \mathfrak{N}_s \cup \{\zeta_s\}$, $\sigma \oplus \iota_s = \sigma$. Notice that a commutative semigroup may have at most one identity. Intuitively, identity valuations represent knowledge that is completely vacuous, i.e., they have no substantive content.

**Marginalization.** We assume that for each nonempty $s \subseteq \mathfrak{X}$, and for each $X \in s$, there is a mapping $\downarrow(s-\{X\}): \mathcal{V}_s \to \mathcal{V}_{s-\{X\}}$, called *marginalization to s-{X}*, that satisfies certain conditions. We call $\sigma^{\downarrow(s-\{X\})}$ the *marginal of $\sigma$ for s-{X}*.

If we regard marginalization as a coarsening of a valuation by deleting variables, then we assume that the order in which the variables are deleted does not matter.

Also we assume that marginalization preserves the coherence of knowledge.

Suppose $\rho \in \mathcal{V}_r$ and $\sigma \in \mathcal{V}_s$. Suppose $X \notin r$, and $X \in s$. Then we assume that
$$(\rho \oplus \sigma)^{\downarrow((r \cup s) - \{X\})} = \rho \oplus (\sigma^{\downarrow(s - \{X\})}).$$

Finally we assume that if $r \subseteq s$, then $\iota_s^{\downarrow r} = \iota_r$.

Intuitively, marginalization corresponds to coarsening of knowledge. If $\sigma$ is a valuation for s representing some knowledge about variables in s, and $X \in s$, then $\sigma^{\downarrow(s-\{X\})}$ represents the knowledge about variables in s–{X} implied by $\sigma$ if we disregard variable X.

The definitions of combination and marginalization make local computation of marginals possible. Suppose $\{\sigma_1, ..., \sigma_m\}$ is a collection of valuations, and suppose $\sigma_i \in \mathcal{V}_{s_i}$. Suppose $\mathfrak{X} = s_1 \cup ... \cup s_m$, and suppose $X \in \mathfrak{X}$. Suppose we wish to compute $(\sigma_1 \oplus ... \oplus \sigma_m)^{\downarrow\{X\}}$. We can do so by successively deleting all variables but X from the collection of valuation $\{\sigma_1, ..., \sigma_m\}$. Each time we delete a variable, we do a fusion operation defined as follows. Consider a set of k valuations $\rho_1, ..., \rho_k$. Suppose $\rho_i \in \mathcal{V}_{r_i}$. Let $\text{Fus}_Y\{\rho_1, ..., \rho_k\}$ denote the collection of valuations after fusing the valuations in the set $\{\rho_1, ..., \rho_k\}$ with respect to variable $Y \in r_1 \cup ... \cup r_k$. Then
$$\text{Fus}_Y\{\rho_1, ..., \rho_k\} = \{\rho^{\downarrow(r - \{Y\})}\} \cup \{\rho_i \mid Y \notin r_i\}$$
where $\rho = \oplus\{\rho_i \mid Y \in r_i\}$, and $r = \cup\{r_i \mid Y \in r_i\}$. After fusion, the set of valuations is changed as follows. All valuations whose domains include Y are combined, and the resulting valuation is marginalized such that Y is eliminated from its domain. The valuations whose domains do not include Y remain unchanged. The following lemma describes an important consequence of the fusion operation.

**Lemma 2.1** [Shenoy 1992a]. Suppose $\{\rho_1, ..., \rho_k\}$ is a collection of valuations such that $\rho_i \in \mathcal{V}_{r_i}$. Let $\mathfrak{X}$ denote $r_1 \cup ... \cup r_k$. Suppose $Y \in \mathfrak{X}$. Then

$$\oplus\text{Fus}_Y\{\rho_1, ..., \rho_k\} = (\rho_1 \oplus ... \oplus \rho_k)^{\downarrow(\mathfrak{X}-\{Y\})}.$$

Next, we define another binary operation called removal. The removal operation is an inverse of the combination operation.

**Removal**. We assume there is a mapping $\ominus: \mathcal{V} \times (\mathfrak{N} \cup \mathcal{Z}) \rightarrow (\mathfrak{N} \cup \mathcal{Z})$, called *removal*, such that if $\sigma \in \mathcal{V}_s$, and $\rho \in \mathfrak{N}_r \cup \mathcal{Z}_r$, then $\sigma \ominus \rho \in \mathfrak{N}_{r \cup s} \cup \mathcal{Z}_{r \cup s}$.

We call $\sigma \ominus \rho$, read as $\sigma$ minus $\rho$, the *valuation resulting after removing $\rho$ from $\sigma$*. Intuitively, $\sigma \ominus \rho$ can be interpreted as follows. If $\sigma$ and $\rho$ represent some knowledge, and if we remove the knowledge represented by $\rho$ from $\sigma$, then $\sigma \ominus \rho$ describes the knowledge that remains.

We assume that the removal operation is an "inverse" of the combination operation in the sense that arithmetic division is inverse of arithmetic multiplication, and in the sense that arithmetic subtraction is inverse of arithmetic multiplication.

**Conditionals**. Suppose $\sigma \in \mathfrak{N}_s$, and suppose a and b are disjoint subsets of s. The valuation $\sigma^{\downarrow(a \cup b)} \ominus \sigma^{\downarrow a}$ for $a \cup b$ plays an important role in the theory of conditional independence. Borrowing terminology from probability theory, we call $\sigma^{\downarrow(a \cup b)} \ominus \sigma^{\downarrow a}$ the *conditional for b given a with respect to $\sigma$*. Let $\sigma(b|a)$ denote $\sigma^{\downarrow(a \cup b)} \ominus \sigma^{\downarrow a}$. We call b the *head* of the domain of $\sigma(b|a)$, and we call a the *tail* of the domain of $\sigma(b|a)$. Also, if $a = \emptyset$, let $\sigma(b)$ denote $\sigma(b|\emptyset)$. The following theorem states some important properties of conditionals.

**Theorem 2.1** [Shenoy 1991b]. Suppose $\sigma \in \mathfrak{N}_s$, and suppose a, b, and c are disjoint subsets of s.
(i). $\sigma(a) = \sigma^{\downarrow a}$.
(ii). $\sigma(a) \oplus \sigma(b|a) = \sigma(a \cup b)$.
(iii). $\sigma(b|a) \oplus \sigma(c|a \cup b) = \sigma(b \cup c|a)$.
(iv). Suppose $b' \subseteq b$. Then $\sigma(b|a)^{\downarrow(a \cup b')} = \sigma(b'|a)$.
(v). $(\sigma(b|a) \oplus \sigma(c|a \cup b))^{\downarrow(a \cup c)} = \sigma(c|a)$
(vi). $\sigma(b|a)^{\downarrow a} = \iota_{\sigma(a)}$, where $\iota_{\sigma(a)}$ is an identity for $\sigma(a)$.
(vii). $\sigma(b|a) \in \mathfrak{N}_{a \cup b}$.

**Conditional Independence**. Suppose $\tau \in \mathfrak{N}_w$, and suppose r, s, and v are disjoint subsets of w. We say *r and s are conditionally independent given v with respect to $\tau$*, written as $r \perp_\tau s \mid v$, if and only if $\tau(r \cup s \cup v) = \alpha_{r \cup v} \oplus \alpha_{s \cup v}$, where $\alpha_{r \cup v} \in \mathcal{V}_{r \cup v}$, and $\alpha_{s \cup v} \in \mathcal{V}_{s \cup v}$.

When it is clear that all conditional independence statements are with respect to $\tau$, we simply say 'r and s are conditionally independent given v' instead of 'r and s are conditionally independent given v with respect to $\tau$,' and use the simpler notation $r \perp s \mid v$ instead of $r \perp_\tau s \mid v$. Also, if $v = \emptyset$, we say 'r and s are independent' instead of 'r and s are conditionally independent given $\emptyset$' and use the simpler notation $r \perp s$ instead of $r \perp s \mid \emptyset$.

Shenoy [1991b] shows that the conditional independence relation generalizes the conditional independence relation in probability theory. In particular, all characterizations of it given by Dawid [1979] (including the graphoid axioms) follow from the above definition.

## 3 VALUATION NETWORKS

In this section, we define a valuation network representation of a VBS and explain how a valuation network encodes conditional independence statements.

A valuation network (VN) consists of a four-tuple $\{\mathfrak{X}, \mathcal{V}, \mathcal{E}, \mathcal{A}\}$ where $\mathcal{E} \subseteq \mathcal{V} \times \mathfrak{X}$, and $\mathcal{A} \subseteq \mathcal{V} \times \mathfrak{X}$. We call the elements of $\mathfrak{X}$ *vertices* and they represent variables. We call the elements of $\mathcal{V}$ *nodes* and they represent valuations. We call the elements of $\mathcal{E}$ *edges*, and they denote either domains of valuations, or tails of domains of conditionals. We call the elements of $\mathcal{A}$ *arcs* and they denote the heads of domains of conditionals. In VNs, vertices are denoted by circles, nodes by diamonds, edges by lines joining the respective nodes and vertices, and arcs by a directed edge pointing to the corresponding vertex. When a VN contains conditionals, we will assume that all conditionals are with respect to valuation $\tau$ obtained by combining all valuations in the network.

**Example 1**. Consider a VBS consisting of variables W, X, Y, and Z, and valuations $\alpha$ for $\{W, X\}$, $\beta$ for $\{X, Y\}$, $\gamma$ for $\{Y, Z\}$, and $\delta$ for $\{W, Z\}$. Figure 1 shows the VN for this VBS. Vertices (variables) are depicted by circles, nodes (valuations) are depicted by diamonds, and edges are depicted by lines. The edges $(\alpha, W)$ and $(\alpha, X)$ incident to node $\alpha$ indicate that $\{W, X\}$ is the domain of $\alpha$.

**Example 2**. Consider a VBS consisting of variables V, W, X, Y, and Z, and conditionals $\alpha$ for $\{V\}$, $\beta$ for $\{W\}$ given $\{V\}$, $\gamma$ for $\{X\}$ given $\{V\}$, $\delta$ for $\{Y\}$ given $\{W, X\}$, and $\varepsilon$ for $\{Z\}$ given $\{Y\}$. Figure 2 shows the VN for this VBS. The arc $(\delta, Y)$ incident to $\delta$ indicates that Y is the head of the domain of $\delta$, and the edges $(\delta, W)$ and $(\delta, X)$ incident to $\delta$ indicate that $\{W, X\}$ is the tail of the domain of $\delta$. Further, if $\tau$ denotes $\alpha \oplus \beta \oplus \gamma \oplus \delta \oplus \varepsilon$, then $\alpha = \tau(V)$, $\beta = \tau(W|V)$, $\gamma = \tau(X|V)$, $\delta = \tau(Y|W,X)$, $\varepsilon = \tau(Z|Y)$. For simplicity, we drop braces around subsets in conditional valuations. Thus we write $\tau(V)$ instead of $\tau(\{V\})$, $\tau(Y|W,X)$ instead of $\tau(\{Y\} \mid \{W, X\})$, etc.

**Example 3**. Consider a VBS consisting of variables $X_1, ..., X_{10}$, and conditionals $\alpha_1$ for $X_1$ given $\emptyset$, $\alpha_2$ for $\{X_2, X_3\}$ given $X_1$, $\alpha_3$ for $X_4$ given $X_2$, $\alpha_4$ for $\{X_5, X_6, X_7\}$ given $X_2$, $\alpha_5$ for $X_8$ given $X_3$, $\alpha_6$ for $X_9$ given $X_5$, and $\alpha_7$ for $X_{10}$ given $\{X_6, X_7\}$. Figure 3 shows the VN for this VBS. If $\tau$ denotes $\alpha_1 \oplus ... \oplus \alpha_7$, then $\alpha_1 = \tau(X_1)$, $\alpha_2 = \tau(X_2, X_3 \mid X_1)$, $\alpha_3 = \tau(X_4 \mid X_2)$, $\alpha_4 = \tau(X_5, X_6, X_7 \mid X_2)$, $\alpha_5 = \tau(X_8 \mid X_3)$, $\alpha_6 = \tau(X_9 \mid X_5)$, $\alpha_7 = \tau(X_{10} \mid X_6, X_7)$.

**Example 4**. Consider a VBS consisting of variables V, W, X, Y and Z, valuations $\alpha$ for (V, W), and $\beta$ for $\{V, X\}$, and conditionals $\gamma$ for Y given $\{W, X\}$, and $\delta$ for Z given X. Figure 4 shows the VN for this VBS. If $\tau$ denotes $\alpha \oplus \beta \oplus \gamma \oplus \delta$, then $\alpha \oplus \beta = \tau(V, W, X)$, $\gamma = \tau(Y \mid W, X)$, and $\delta = \tau(Z \mid X)$.

**Fusion in Valuation Networks**. Next, we will illustrate fusion in VNs. Consider the VBS described in Example 1. If we fuse the valuations in the set $\{\alpha, \beta, \gamma, \delta\}$ with respect to X, we get $\text{Fus}_X\{\alpha, \beta, \gamma, \delta\} =$

$\{(\alpha\oplus\beta)^{\downarrow\{Y,Z\}}, \gamma, \delta\}$.
Figure 5 illustrates this fusion operation. Lemma 2.1 tells us that
$(\alpha\oplus\beta)^{\downarrow\{Y,Z\}}\oplus\gamma\oplus\delta = (\alpha\oplus\beta\oplus\gamma\oplus\delta)^{\downarrow\{Y,Z,W\}}$.

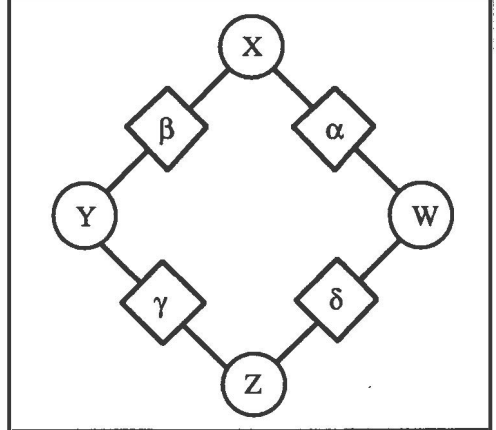

Fig. 1. The VN for the VBS of Example 1.

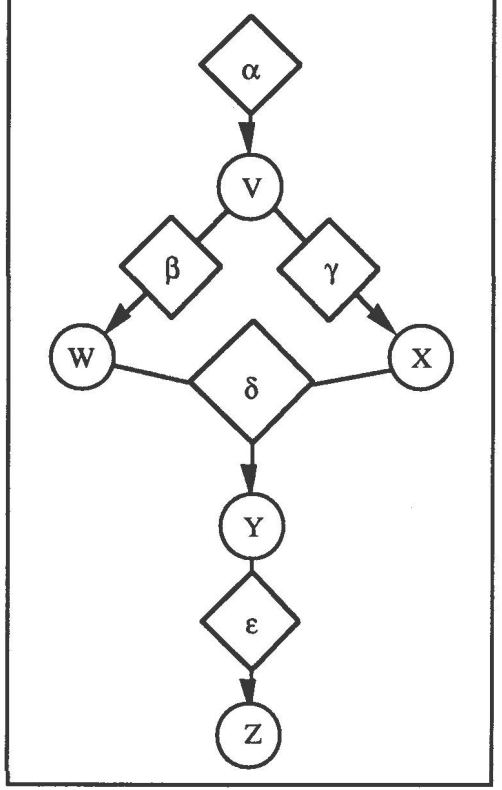

Fig. 2. The VN for the VBS of Example 2.

Next, we will illustrate fusion in VNs when we have conditionals. The various statements of Theorem 2.1 are useful here. Consider a VBS consisting of four variables W, X, Y, and Z, and three conditionals, $\alpha$ for W given $\emptyset$, $\beta$ for X given W, and $\gamma$ for $\{Y, Z\}$ given X. Figure 6 shows the corresponding VN. Suppose $\tau = \alpha\oplus\beta\oplus\gamma$. Then $\alpha = \tau(W)$, $\beta = \tau(X\mid W)$, and $\gamma = \tau(Y, Z\mid X)$. After fusion with respect to X, we have two conditionals $\alpha = \tau(W)$, and $(\beta\oplus\gamma)^{\downarrow\{W, Y, Z\}} = \tau(Y, Z\mid W)$. This result is justified by statement (v) of Theorem 2.1. After further fusion with respect to Z, we only have two conditionals $\alpha = \tau(X)$, and $(\beta\oplus\gamma)^{\downarrow\{W, Y\}} = \tau(Y\mid W)$. This result is justified by statement (iv) of Theorem 2.1. Finally after further fusion with respect to Y, we have only one conditional $\alpha = \tau(W)$. This is because from statement (vi) of Theorem 2.1, $\tau(Y\mid W)^{\downarrow\{W\}}$ is an identity for $\tau(W)$, and this identity can be absorbed in any conditional that has W in the head of its domain.

**Conditional Independence in Valuation Networks.** How is conditional independence encoded in VNs? Let us examine the definition of conditional independence graphically. Suppose r, s, and v are disjoint subsets of variables, and suppose $\tau$ is a normal valuation for $r\cup s\cup v$. Our definition of conditional independence states that $\tau = \alpha_{r\cup v}\oplus\alpha_{s\cup v}$ iff $r\perp_\tau s\mid v$, where $\alpha_{r\cup v}\in\mathcal{V}_{r\cup v}$, and $\alpha_{s\cup v}\in\mathcal{V}_{s\cup v}$. Suppose $\tau = \alpha_{r\cup v}\oplus\alpha_{s\cup v}$ is a normal valuation for $r\cup s\cup v$, where $\alpha_{r\cup v}\in\mathcal{V}_{r\cup v}$, and $\alpha_{s\cup v}\in\mathcal{V}_{s\cup v}$. Figure 7 shows the VN representation of this situation. Notice that all paths from a variable in r to a variable in s go through a variable in v, i.e., v is a cut-set separating r from s.

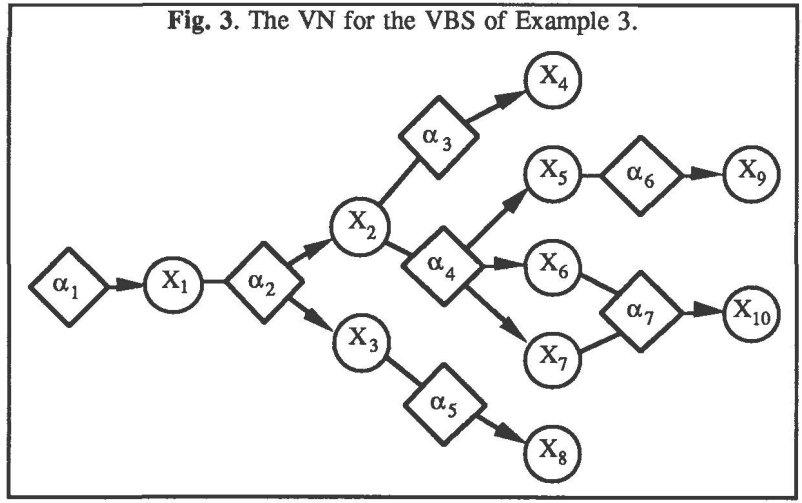

Fig. 3. The VN for the VBS of Example 3.

Suppose r, s, and v are disjoint subsets of w. Suppose $\tau\in\mathcal{V}_w$. Consider the VN representation of $\tau(r\cup s\cup v)$ *after* marginalizing all variables in $w-(r\cup s\cup v)$ out of $\tau$. Suppose v is a cut-set separating r and s. Then there is no valuation that contains a variable in r and a variable in s. Consider all valuations whose domain includes a variable in r. Let $\rho$ denote the combination of these valuations. Notice that the domain of $\rho$ does not contain a variable in s. Now consider all valuations whose domain includes a variable in s. Let $\sigma$ denote the combination of these valuations. Notice that the domain of $\sigma$ does not include a variable in r. Finally let $\theta$ denote the combination of all valu-ations not included in either ρ or σ. Clearly, the domain of θ does not contain variables in either r or s. Since $\tau(r\cup s\cup v) = \rho\oplus\sigma\oplus\theta$, it follows from the definition of conditional independence that $r \perp_\tau s \mid v$.

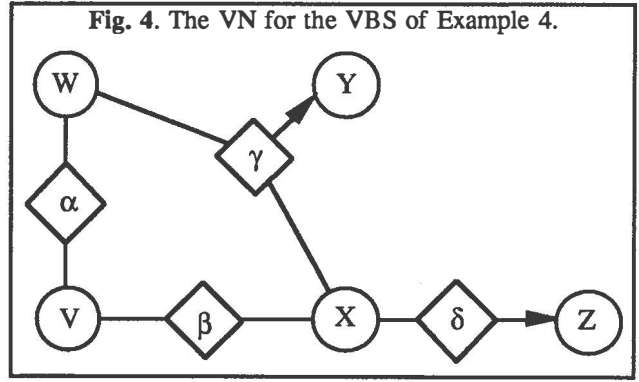

Fig. 4. The VN for the VBS of Example 4.

To summarize, suppose we are given a VN representation of $\tau \in \mathfrak{N}_w$. Suppose $v$ is a cut-set separating r and s in the marginalized network for variables in $r\cup s\cup v$. Then $r\perp_\tau s\mid v$.

## 4 COMPARISON

In this section, we briefly compare VNs with UGs, DAGs, DBGs, and RCGs. We start with UGs.

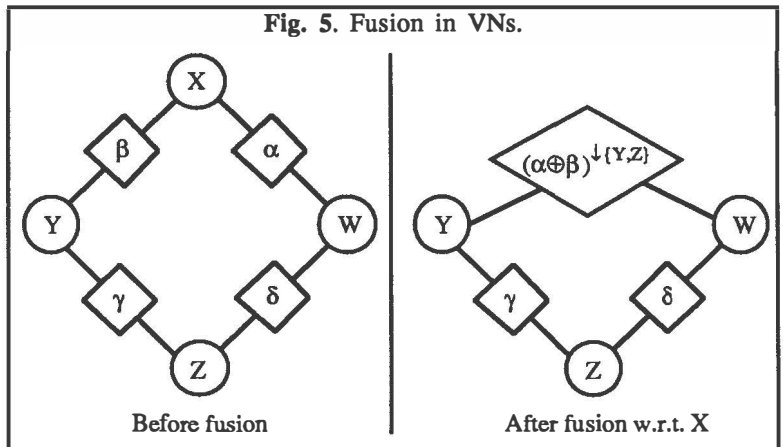

Fig. 5. Fusion in VNs.

In UGs, the cliques of the graph (maximal completely connected vertices) denote the factors of the joint valuation. For example, consider the UG shown in Figure 8. This graph has 4 cliques, {W, X}, {X, Y}, {Y, Z}, and {Z, X}. This undirected graph models a joint probability distribution for {W, X, Y, Z} that factors (multiplicatively) into 4 components, α with domain {W, X}, β with domain {X, Y}, γ with domain {Y, Z}, and δ with domain {Z, W}. The VN representation of this distribution is also shown in Figure 8. Notice that for this distribution, $\{X\}\perp\{Z\}\mid\{Y,W\}$, and $\{Y\}\perp\{W\}\mid\{X,Z\}$.

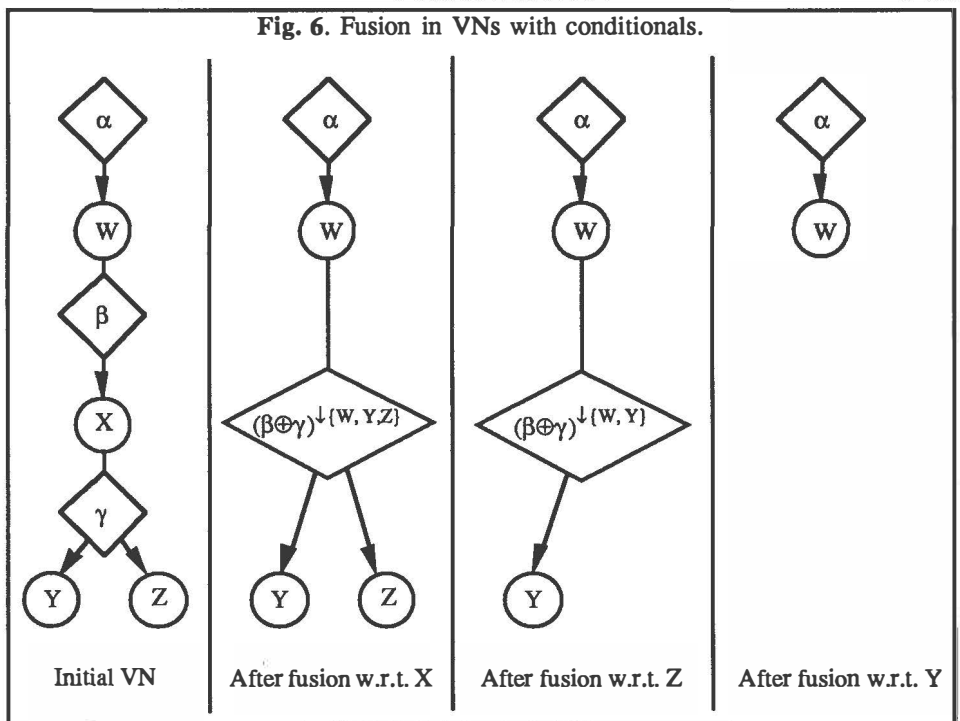

Fig. 6. Fusion in VNs with conditionals.

Next, we consider DAGs. A DAG model of a probability distribution consists of an ordering of the variables, and a conditional for each variable given a subset of the variables that precede it in the given ordering. Figure 9 shows an example of a DAG with 5 variables. An ordering consistent with this DAG is VWXYZ. The DAG implies we have a conditional for V given ∅, a conditional for W given V, a conditional for X given V, a conditional for Y given {W, X}, and a conditional for Z given Y. The VN representation of the DAG model is also shown in Figure 9. Suppose τ denotes the joint probability distribution. Then $\alpha = \tau(V)$, $\beta = \tau(W\mid V)$, $\gamma = \tau(X\mid V)$, $\delta = \tau(Y\mid W, X)$, and $\varepsilon = \tau(Z\mid Y)$.

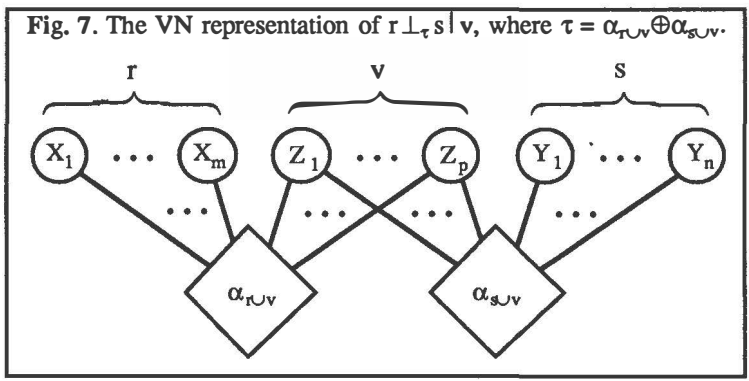

Fig. 7. The VN representation of $r\perp_\tau s\mid v$, where $\tau = \alpha_{r\cup v}\oplus\alpha_{s\cup v}$.

In the DAG of Figure 9, using Pearl's definition of d-separation, we cannot conclude, for example, that $W\perp_\tau X\mid \{V, Z\}$. However, we can conclude that $W\perp_\tau X\mid V$. We can draw the same conclusion using separation in VNs. If we fuse the VN with respect to Y, then {V, Z} is not a cut-set separating W and X. Therefore we cannot conclude that $W\perp_\tau X\mid\{V, Z\}$. If we further fuse the VN with respect to Z, then V is a cut-set separating W and X. Therefore, $W \perp_\pi X \mid V$.

The technique we have proposed for checking for conditional independence in VNs is an alternative to the d-separation method proposed by Pearl [1988] for DAGs. Whether we have conditionals or not, checking a conditional independence statement in a VN is a matter of first fusing the VN to remove variables not in the conditional independence statement and then checking for separation in the fused VN. The information about conditionals is used in the fusion operation.

Lauritzen et al. [1990] describes yet another method for checking for conditional independence in DAGs. Their method consists of converting a DAG to an equivalent UG and then checking for conditional independence in the UG using separation. In short, their method consists of examining a subgraph of the DAG (after eliminating the variables that succeed all variables in the conditional independence statement in an ordering consistent with the DAG), moralizing the graph, dropping directions, and then checking for separation.

Next, let us compare VNs and DBGs. DBGs are defined in [Shafer 1993a]. A DBG includes a partition of the set of all variables. Each element of the partition is called a balloon. Non-singleton balloons are shown as ellipses encircling the corresponding variables. Each balloon has a set of variables as its parents. The parents of a balloon are shown by directed arcs pointing to the balloon. A DBG is acyclic in the same sense that DAGs are acyclic. A DBG implies a probability model consisting of a conditional for each balloon given its parents. A DAG may be considered as a DBG in which each balloon is a singleton subset.

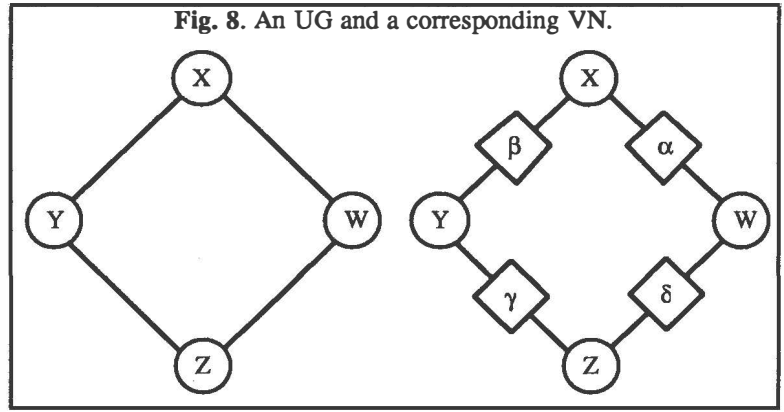

Fig. 8. An UG and a corresponding VN.

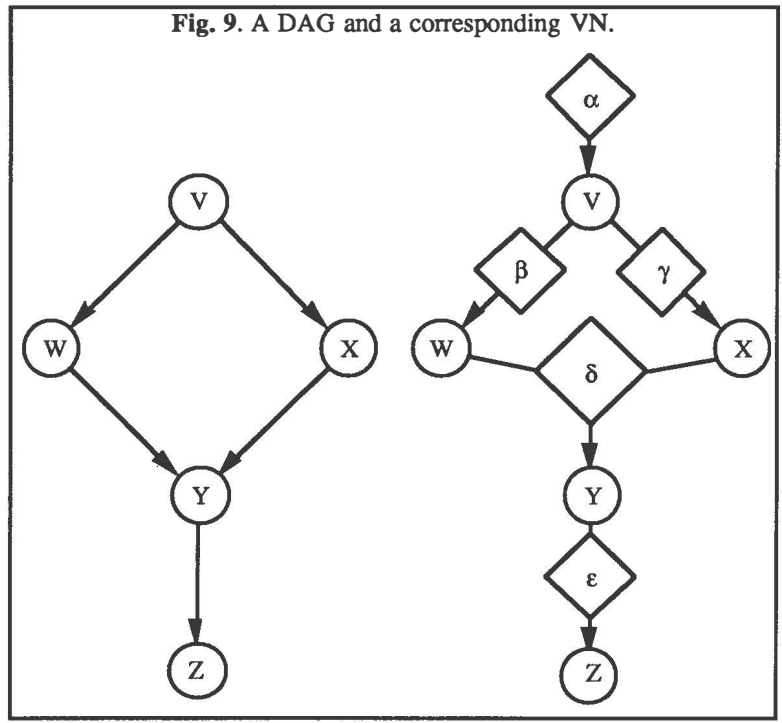

Fig. 9. A DAG and a corresponding VN.

Figure 10 shows a DBG with 10 variables, $X_1$, ..., $X_{10}$. There are two non-singleton balloons, $\{X_2, X_3\}$, and $\{X_5, X_6, X_7\}$. All other balloons are singleton subsets. The DBG of Figure 10 implies a conditional for $X_1$ given $\emptyset$, a conditional for $\{X_2, X_3\}$ given $X_1$, a conditional for $X_4$ given $X_2$, a conditional for $\{X_5, X_6, X_7\}$ given $X_2$, a conditional for $X_8$ given $X_3$, a conditional for $X_9$ given $X_5$, and a conditional for $X_{10}$ given $\{X_6, X_7\}$. The corresponding VN is also shown in Figure 10.

The conditional independence theory of DBGs is described in [Shafer 1993b], and is analogous to the conditional independence theory of DAGs. In the DBG and VN of Figure 10, we have, for example, $\{X_5, X_6, X_7\} \perp \{X_1, X_3, X_4\} \mid \{X_2\}$.

Finally, we compare VNs to RCGs. RCGs are defined in [Kiiveri, Speed, and Carlin 1984]. A RCG consists of two kinds of vertices (variables)—exogenous and endogenous, and two kinds of edges—undirected and directed. An undi-rected edge always connects two exogenous variables, and a directed edge always points to an endogenous variable. RCGs generalize DAGs in the sense that a DAG is a RCG with at most one exogenous variable.

Figure 11 shows a RCG with five variables, V, W, X, Y, Z. Variables V, W, and X are exogenous, and variables Y and Z are endogenous. The cliques $\{V, W\}$ and $\{V, X\}$ imply valuations for $\{V, W\}$ and $\{V, X\}$ respectively. The directed edges pointing to Y imply a conditional for Y given $\{W, X\}$, and the directed edge pointing to Z implies a conditional for Z given X. The corresponding VN is also shown in Figure 11.

Conditional independence properties of RCG are given in [Kiiveri, Speed and Carlin 1984]. Briefly, if we look at the subgraph of a RCG restricted to the exogenous variables, the subgraph is an UG and its conditional independence properties are the same as those given by the UG models. On the other hand, the conditional independence relation in the complete RCG is given by the d-separation relation of DAGs. Since the basis of the conditional independence relations in RCGs is the underlying factorization and the additional information about conditionals, and since this information is encoded in VNs, a corresponding VN encodes the same conditional independence relation as a RCG. For example, in the RCG and VN of Figure 11, we have $\{W\} \perp \{X\} \mid \{V\}$, and $\{Z\} \perp \{V, W, Y\} \mid \{X\}$.

## 5 CONCLUSION

We have described valuation networks and how they encode conditional independence. Given a valuation network, $r \perp_t s \mid v$ if $v$ is a cut-set separating $r$ from $s$ in the marginalized valuation network for $r \cup s \cup v$. We have compared valuation networks to undirected graphs, directed acyclic graphs, directed balloon graphs, and recursive causal graphs. All probability models encoded by one of these graphs can be represented by corresponding valuation networks.

Factorization is fundamental to conditional independence. The power of the valuation network representation arises from the fact that it represents factorization explicitly. Also notice that valuation networks encode conditional independence not only in probabilistic models, but also in all uncertainty theories that fit in the VBS framework. This includes Dempster-Shafer's belief-function theory, Spohn's epistemic belief theory, and Zadeh's possibility theory [Shenoy 1991b].

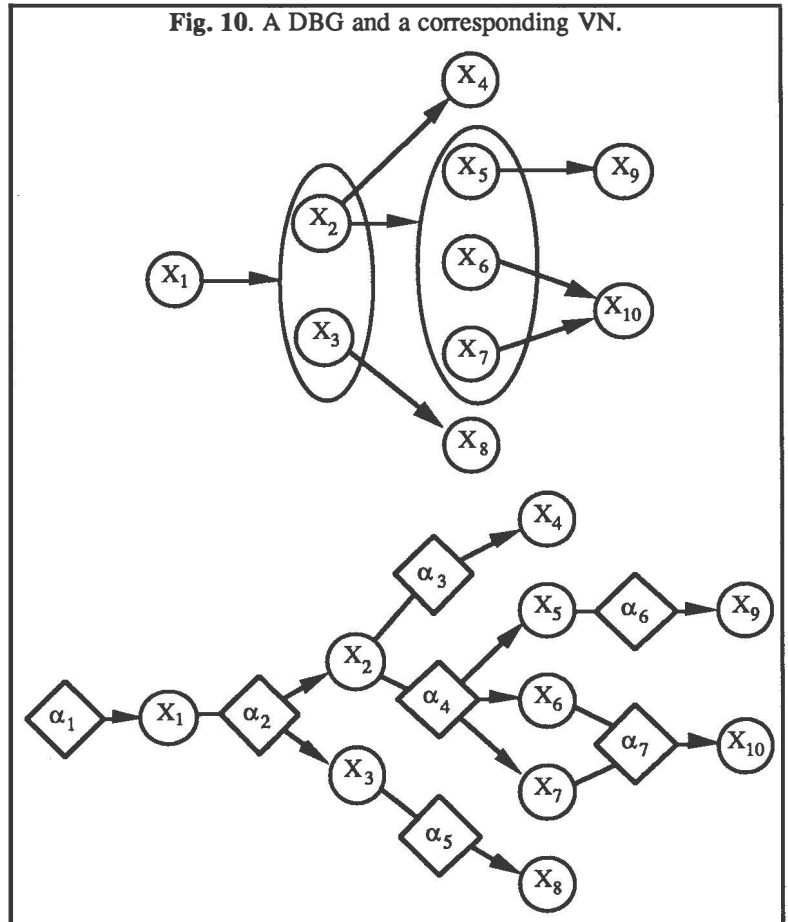

Fig. 10. A DBG and a corresponding VN.

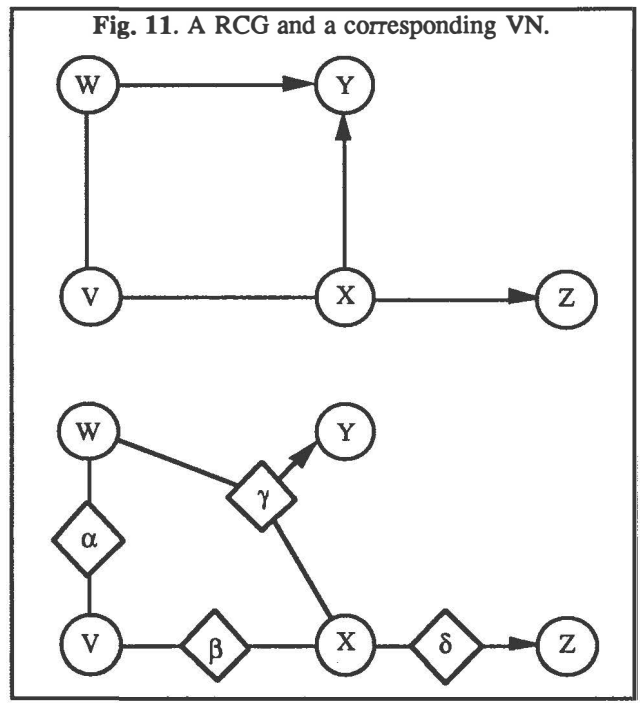

Fig. 11. A RCG and a corresponding VN.

### Acknowledgments

This work is based upon work supported in part by the National Science Foundation under Grant No. SES-9213558, and in part by the General Research Fund of the University of Kansas under Grant No. 3605-XX-0038.